\documentclass[10pt,twocolumn,letterpaper]{article}

\usepackage{icb}
\usepackage{times}
\usepackage{epsfig}
\usepackage{graphicx}
\usepackage{amsmath}
\usepackage{amssymb}
\usepackage{multirow}

\icbfinalcopy 

\ificbfinal\fi
\begin{document}

\title{Deep Cross Polarimetric Thermal-to-visible Face Recognition}

\author{Seyed Mehdi Iranmanesh, Ali Dabouei, Hadi Kazemi, Nasser M. Nasrabadi\\
West Virginia University\\
{\tt\small \{seiranmanesh, ad0046, hakazemi\}@mix.wvu.edu, \{nasser.nasrabadi\}@mail.wvu.edu}}
\maketitle
\thispagestyle{empty}

\begin{abstract}
In this paper, we present a deep coupled learning framework to address the problem of matching polarimetric thermal face photos against a gallery of visible faces. Polarization state information of thermal faces provides the missing textural and geometrics details in the thermal face imagery which exist in visible spectrum. we propose a coupled deep neural network architecture which leverages relatively large visible and thermal datasets to overcome the problem of overfitting and eventually we train it by a polarimetric thermal face dataset which is the first of its kind. The proposed architecture is able to make full use of the polarimetric thermal information to train a deep model compared to the conventional shallow thermal-to-visible face recognition methods. Proposed coupled deep neural  network also finds global discriminative features in a nonlinear embedding space to relate the polarimetric thermal faces to their corresponding visible faces. The results show the superiority of our method compared to the state-of-the-art models in cross thermal-to-visible face recognition algorithms. 
\end{abstract}

\section{Introduction}
 
 Face recognition has been one of the most challenging areas of research in biometrics and computer vision. Many face recognition algorithms are dedicated to address pose and illumination problems for visible face images. In recent years, there has been significant amount of research in Heterogeneous Face Recognition (HFR)~\cite{1}. The main issue in HFR is to match the visible face image to a face image that has been captured in other sensing modalities such as infrared spectrum. Infrared images are divided into two major categories of reflection and emission. The reflection category, which consists of near infrared (NIR) and shortwave infrared (SWIR) bands, is more similar to the visible imagery and it is more informative about the facial details of the face. Due to this reflective phenomenology of the NIR and SWIR, there has been success on NIR-to-visible face recognition~\cite{2,3} and SWIR-to-visible face recognition~\cite{4,5} to some extent. Klare et al.~\cite{6} used kernel similarities for a set of training subjects as features. A novel transductive subspace learning method was proposed in~\cite{7} for domain invariant feature extraction for VIS-NIR matching problem. In~\cite{8} the authors used Restricted Boltzmann Machine (RBM) to learn a shared representation of features locally and consequently removed the heterogeneity around each facial point. They applied PCA to the local features extracted from RBM to get the high level features. Juefei-Xu et al.~\cite{9} used a dictionary learning approach to reconstruct images between visible and NIR domains. A common weakness of the mentioned methods is that they are not using a deep global features of face images, which has been shown to have better results in face recognition problems~\cite{42}. 
 
 \begin{figure}
 	\begin{center}
 		\includegraphics[width=1\linewidth]{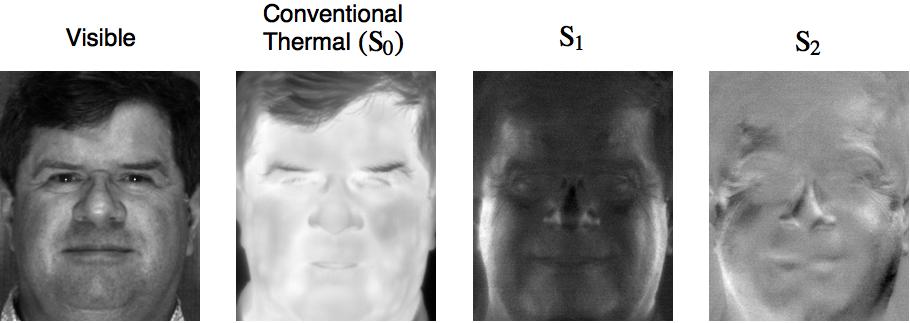}
 		
 	\end{center}
 	\caption{Visible spectrum and its corresponding conventional thermal, and polarimetric thermal images of a subject.}
 	\label{fig:figure1}
 \end{figure}

 Compared with the reflection category, the emission group which contains the midwave infrared (MWIR) and longwave infrared (LWIR) bands is less informative~\cite{10}. Due to the inherent phenomenology of thermal imaging which is significantly different from visible imagery, matching a thermal face against a gallery of visible faces becomes a challenging task. However, thermal-to-visible recognition is highly demanding as the thermal data is illumination invariant and useful for face recognition in the cover of darkness. In recent years, there has been a growing research on thermal-to-visible face recognition ~\cite{11,12,6,14} and thermal-to-visible detection~\cite{46}. Furthermore, via an emerging technology, the polarization state information of thermal emission has been exploited to provide the geometrical and textural details of a face. This information which is not available in the conventional intensity-based thermal imaging~\cite{15}, is used to improve the cross thermal-to-visible face recognition~\cite{16,17,15}. Figure~\ref{fig:figure1} shows a visible image and its corresponding conventional thermal and polarimetric thermal images. In cross-spectrum face recognition a thermal probe is matched against a gallery of visible faces, corresponding to the real-world scenario~\cite{19}. Researchers have also investigated a variety of approaches to exploit the polarimetric thermal images in LWIR to improve cross-spectrum face recognition~\cite{15,16,22,23}.

 \begin{figure}
 	\begin{center}
 		\includegraphics[width=1\linewidth]{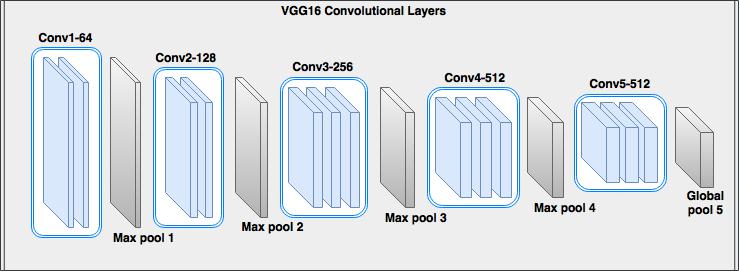}
 		
 	\end{center}
 	\caption{shows the first 13 convolutional layers of a VGG-16 architecture.}
 	\label{fig:figure2}
 \end{figure}
 
Recently, almost all the state-of-the-art techniques in face recognition have applied Deep Convolutional Neural Networks (DCNN) trained on extremely huge datasets to construct a compact discriminative feature representation. This approach also has been applied in other cross-modal applications such as~\cite{45,46} to find a representative embedding space. In~\cite{24}, the authors who were the pioneers in training a Deep Neural Network (DNN) for face recognition, trained a network on a private dataset containing 4.4 million labeled images of 4030 different subjects. They finally fine-tuned their network with a Siamese network~\cite{25} for the face verification task. They also extended their work with an expanded dataset which contains 500 million images related to 10 million subjects. Sun et al.~\cite{26,27,28,29} studied a different deep neural network architecture including a joint verification-identification loss function and Bayesian metrics in their works. They used two different datasets, namely, CelebFaces~\cite{26} (202,599 images of 10,177 different subjects) and WDRef~\cite{31} (99,773 images of 2995 subjects) to train their deep networks. A triplet loss function is utilized in~\cite{32} to train a feature embedding space. Schroff et al.~\cite{43} also trained a deep network using 200 million images of 8 million different subjects. This network has the best performance on Labeled Faces in the Wild (LFW)~\cite{33}, which is a standard unconstrained face recognition benchmark.     
\begin{figure*}[t]
	\begin{center}
		\includegraphics[width=1\linewidth]{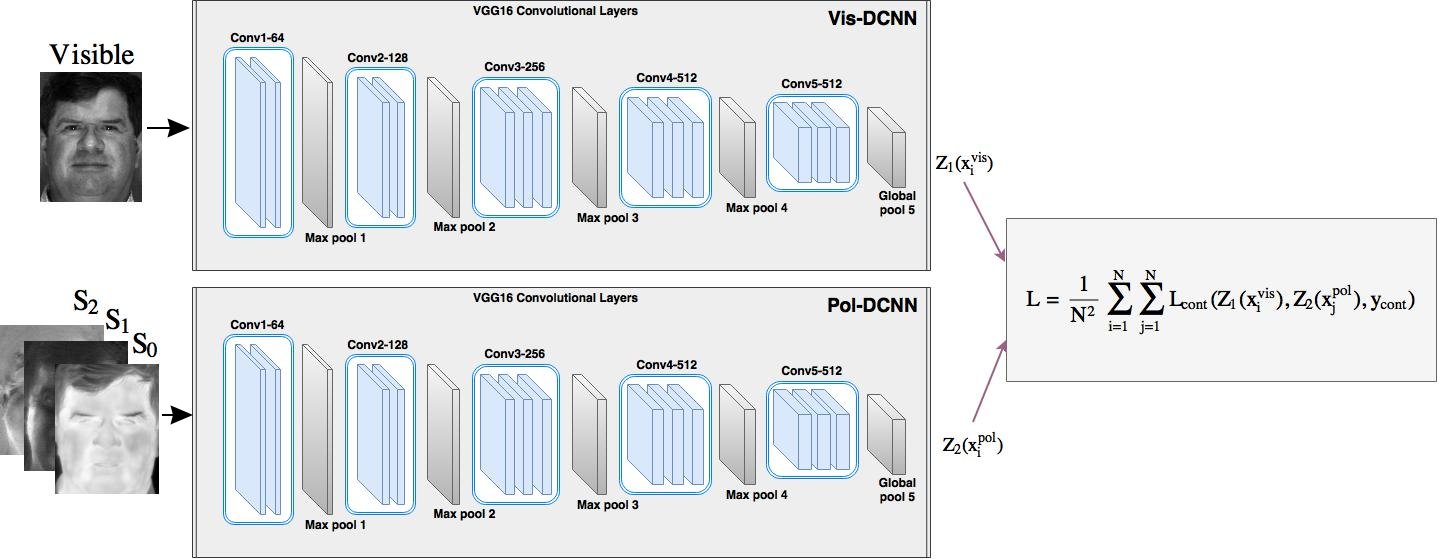}
		
	\end{center}
	\caption{Proposed network using two convolutional networks (Vis-DCNN and Pol-DCNN) coupled by contrastive loss function.}
	\label{figure3coupled}
\end{figure*}

Motivated by recent advances in face recognition algorithms using DCNN, in this paper we propose a novel Coupled Deep Convolutional Neural Network (CpDCNN) for polarimetric thermal-to-visible face recognition, which utilizes both the thermal and polarization state information to enhance the performance of a cross-spectrum face recognition system. In~\cite{42}, they used a coupled architecture for their face recognition system. However, they evaluated their framework only for near infrared which is very close to visible and they had a huge number of data for both modalities. Here, we evaluate the proposed algorithm on a thermal dataset which is more challenging due to the difference between the modalities and the small size of dataset. Scarcity of dataset is an active area of research which has been investigated in other applications~\cite{48}. We applied our framework on thermal polarimetric dataset which contains face images that has been taken at three different ranges and with different face expressions. We compare our proposed framework against several different state-of-the-art techniques in the literature such as deep perceptual mapping technique (DPM)~\cite{34}, a coupled neural network (CpNN)~\cite{35}, and a partial least squares (PLS)~\cite{14}, PLS$\circ$DPM and PLS$\circ$CpNN~\cite{19}. Our results show that our proposed deep method could fuse polarimetric and thermal features in a way to outperform the conventional methods and enhance the performance of a thermal-to-visible face recognition system.  

\section{Background}

The Stokes parameters $S_0$, $S_1$, $S_2$, and $S_3$ are the thermal polarization state information that are captured from an object. The polarimetric measurement is done using a series of linear and circular polarizers. The four mentioned Stokes parameters which completely define the polarization states are: 

\begin{equation}
S_0 = I_0^{\circ} +I_{90}^{\circ} \;,
\label{eq-1}
\end{equation}
\begin{equation}
S_1 = I_0^{\circ} -I_{90}^{\circ} \;,
\label{eq-1}
\end{equation}
\begin{equation}
S_2 = I_{45}^{\circ} +I_{-45}^{\circ} \;,
\label{eq-1}
\end{equation}
\begin{equation}
S_3 = I_R^{\circ} +I_L^{\circ} \;,
\label{eq-1}
\end{equation}

\noindent where $I_0^{\circ}$, $I_{90}^{\circ}$, $ I_{45}^{\circ}$, and $I_{-45}^{\circ}$ represent the measured intensity of the light after passing through a linear polarizer with angle of $0^{\circ} $, ${90}^{\circ}$, $ {45}^{\circ}$, and $ {-45}^{\circ}$  related to horizontal axes, respectively. $I_R$ and $I_L$ are the intensity of the light after passing through right and left circularly polarization filters. Since there is no artificial illumination in passive imaging, there is almost no circularly polarized information in LWIR or MWIR spectrum. Therefore, $S_3$ is considered to be zero for most of the applications. The linear combination of the remaining Stokes, namely, the Degree of Linear Polarization (DoLP), is computed as: 
\begin{equation}
DoLP = \dfrac{\sqrt{S_1^2 + S_2^2}}{S_0}  \;.
\label{eq-1}
\end{equation}

\section{Deep cross polarimetric thermal-to-visible face recognition}

In this paper, we used a VGG-16 like network~\cite{37} in our cross-spectrum recognition framework. The VGG-16 neural network comprised of five major convolutional components which are connected in series (see Figure~\ref{fig:figure2}). The first two components, $Conv1-64$ and $Conv2-128$ consists of the following layers: a convolutional layer, a rectified linear unit layer, a second convolutional layer, a second rectified linear unit layer, and a max pooling layer. The remaining three components contain one additional convolutional layer and a rectifier linear unit layer. The only exception is in the last component, where global pooling was used instead of the max pooling to reduce the number of parameters.    

The final objective of the proposed model is identification of the polarimetric thermal images of the probe faces while we do not have access to them during the training phase. For this reason, we coupled two VGG-16 like networks one dedicated to the visible spectrum (Vis-DCNN) and the other one to the polarimetric thermal (Pol-DCNN). Each DCNN performs a non-linear transformation of the input space. The ultimate goal of our proposed CpDCNN is to find the global deep features representing the relationship between polarimetric thermal face images and their corresponding visible ones. In other to find the common embedding space between these two different domains we coupled two VGG-16 structured networks (Vis-DCNN and Pol-DCNN) via a contrastive loss function~\cite{25}, which has been used in other applications such as domain adaptation~\cite{47}. This function ($\ell_{cont}$) pulls the genuine pairs (i.e., a face visible image with its own corresponding polarimetric face images) toward each other into a common latent feature subspace and push the impostor pairs (i.e., a face visible image with another subject's polarimetric thermal face image) apart from each other (see Figure~\ref{figure3coupled}). 
Despite the use of a contrastive loss function in the Siamese network~\cite{25}, due to the heterogeneous nature of the visible and polarimetric thermal images, we cannot use the weight sharing of the Siamese network here. Similar to~\cite{25}, the contrastive loss is of the form of:
\begin{align}
\ell_{cont}&(z_1(x_i^{vis}),z_2(x_j^{pol}),y_{cont})=\\ \nonumber &(1-y_{cont})L_{gen}(D(z_1(x_i^{vis}),z_2(x_j^{pol}))+\\ \nonumber & y_{cont}L_{imp}(D(z_1(x_i^{vis}),z_2(x_j^{pol})) \; ,
\label{eq-sda-separation}
\end{align}
where $x_i^{vis}$ is the input for the Vis-DCNN (i.e., visible face image), and $x_j^{pol}$ is the input for the Pol-DCNN (i.e., polarimetric face images). $y_{cont}$ is a binary label, $L_{gen}$ and $L_{imp}$ represent the partial loss functions for the genuine and impostor pairs, respectively, and $D(z_1(x_i^ {vis}),z_2(x_j^{pol}))$ indicates the Euclidean distance between the embedded data in the common feature subspace. The binary label, $y_{cont}$, is assigned a value of 0 when both modalities, i.e., visible and polarimetric, form a genuine pair, or, equivalently, the inputs are from the same class. On the contrary, when the inputs are from different classes, which means they form an impostor pair, $y_{cont}$ is equal to 1. In addition, $L_{gen}$ and $L_{imp}$ are defined as follows: 
\begin{equation}
\begin{split}
L_{gen}(D(z_1(x_i^{vis}),z_2(x_j^{pol})))= &\dfrac{1}{2}  D(z_1(x_i^{vis}),z_2(x_j^{pol}))^2 \\
& \text{for} \quad\quad y_i=y_j  \;.
\label{eq-sda-separation}
\end{split}
\end{equation}
\begin{align}
& L_{imp}(D(z_1(x_i^{vis}), z_2(x_j^{pol})))=\\ \nonumber
& \dfrac{1}{2} \:  max(0,m-D(z_1(x_i^{vis}),z_2(x_j^{pol})))^2\quad   \text{for} \quad y_i \neq y_j\; .
\label{eq-sda-separation} 
\end{align}

Therefore, the loss function can be written as:
\begin{equation}
\begin{split}
L &= 1/N^2 \displaystyle\sum_{i=1}^{N}\displaystyle\sum_{j=1}^{N} \ell_{cont}(z_1(x_i^{vis}), z_2(x_j^{pol}), y_{cont}) \;,
\label{eq-5-cont}
\end{split}
\end{equation}

\noindent where $N$ is the number of samples and $z_1 $ and $z_2 $ are the deep convolutional neural network based embedding functions, which transform $x_i^{vis}$ and $ x_j^{pol}$ into a common latent embedding subspace, respectively. It should be noted that the contrastive loss function~\cite{25} inherently considers the subjects' labels inherently. Therefore, it has the ability to find a discriminative embedding space by employing the data labels in contrast to some other metrics such as Euclidean distance. This discriminative embedding space would be useful in identifying a polarimetric probe photo against a gallery of visible photos.

\begin{figure}
	\begin{center}
		\includegraphics[width=0.70\linewidth]{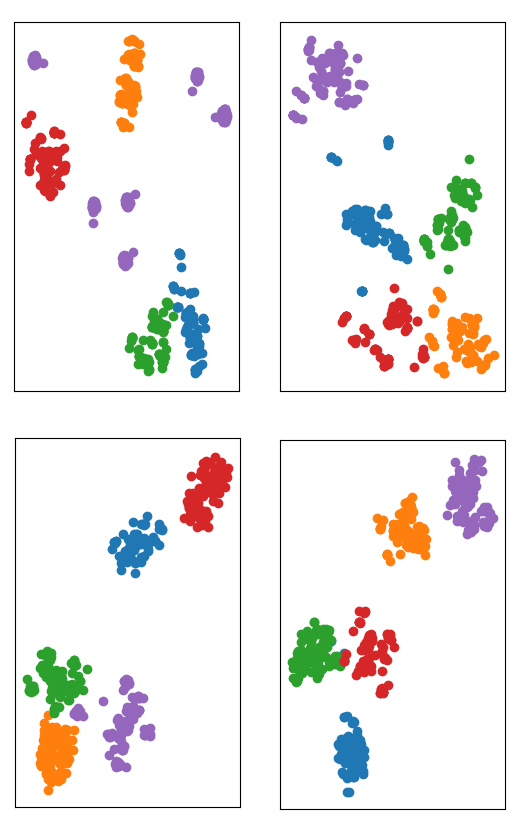}
		
	\end{center}
	\caption{Common embedding subspace of Pol-DCNN in our proposed CpDCNN network for four different groups of 5 subjects.}
	\label{fig:tsne}
\end{figure}

To visualize the latent embedding subspace of the Pol-DCNN network for the polarization dataset trained according to (\ref{eq-5-cont}), we reduced its dimension using Principal Component Analysis (PCA)~\cite{40}. Afterwards, the t-Distributed Stochastic Neighbor Embedding~\cite{41} is employed to project the transformed common features into two dimensions for visualization. The final two dimensional embedding features are depicted in Figure~\ref{fig:tsne}. The plots related to the Pol-DCNN network which is trained in our CpDCNN framework using (\ref{eq-5-cont}) shows a  discriminative embedding space. This emphasizes the effectiveness of our proposed coupling network. 

During the testing phase, for a given test probe $x_t^{pol}$, the proposed CpDCNN is used to transform it to the common latent embedding domain, $z_2(x_t^{pol})$. In fact, after training our deep coupled network model, it has the ability to transform the visible and polarimetric face images into a common discriminative embedding space. Therefore, the galley of the face visible images is transformed to the mentioned embedding space. Eventually, the identification of face polarimetric image is done, by calculating the minimum Euclidean distance between the transformed polarimetric prob and visible gallery images as follows:     

\begin{equation}
\begin{split}
 x_t^{vis} = \underset{x_i^ {vis}}{\operatorname{argmin}}\quad D(z_1(x_i^{vis}),z_2(x_t^{pol}))  \;,
\label{eq-6}
\end{split}
\end{equation}

\noindent where $ x_t^{pol}$ is the polarimetric probe face image and $x_t^{vis}$ is the selected matching visible face image within the gallery of face images.

\section{Experiments and results}

Experiment is performed using three main datasets, such that the first two datasets (CMU Multi-PIE and Notre Dame LWIR Face datasets) were used for initializing and pre-training a CpDCNN network, and the third one (Polarimetric Thermal Face dataset) which is the main objective of the network is used for the final fine-tuning and testing the network.  

\noindent{\bf CMU Multi-PIE} dataset~\cite{38} consists of 337 identities, from 15 different view points and 19 illumination conditions. This dataset contains 750,000 images collected in four different sessions. We used the frontal view face images to train our DCNNs to obtain the initial weights for Vis-DCNN and Pol-DCNN. 

\noindent{\bf Notre Dame LWIR Face} dataset~\cite{39} contains LWIR and visible images related to 159 subjects with different variation in lighting, expression and time lapse. 

\noindent{\bf Polarimethric Thermal Face} dataset~\cite{19} comprises polrimetric LWIR face images and their corresponding visible spectrum related to 60 subjects. Data was collected at three different distances: Range 1 (2.5 m), Range 2 (5 m), and Range 3 (7.5 m). At each range two different conditions, including baseline and expression are considered. In the baseline condition the subject is asked to keep a neutral expression looking at the polarimetric thermal sensor. On the other hand, in the expression condition the subject is asked to count out numerically from one upwards which results in different expressions in the mouth and to the eyes and consequently different variations in the facial imagery. Each subject has 16 images of visible and 16 polarimetric LWIR images in which four images are related to the baseline condition and the remaining 12 images are related to the expression condition.

The network for visible face images (Vis-DCNN) composed of 13 convolutional layers of VGG-16 (Figure~\ref{fig:figure2}), pre-trained on the Imagenet dataset~\cite{44}, followed by three fully connected layers with output sizes of 512, 512, and 337. All the convolutional and fully connected layers, except the last fully connected layer, are equipped with the {\it Relu} activation function. For the sake of classification, {\it softmax} activation is used for the last fully connected layer. After training Vis-DCNN with CMU-Multi PIE dataset, the polarimetric thermal network (Pol-DCNN) was initialized with the Vis-DCNN weights. Afterwards, both networks (Vis-DCNN and Pol-DCNN) are coupled via contrastive loss function which is applied on the output of the global pooling from each network to construct the CpDCNN framework. Since the final goal of the CpDCNN network was not the classification, we exclude the last three fully connected layers from the Vis-DCNN and Pol-DCNN. This also helps to overcome the problem of overfitting for this dataset due to the reduced number of trainable parameters. The proposed network structure (CpDCNN) is depicted in Figure~\ref{figure3coupled}.

After initializing each network (Vis-DCNN and Pol-DCNN) with the CMU Multi-PIE, the CpDCNN network was fine-tuned by the Notre Dame face dataset. To increase the correlation between the two modalities of visible and thermal, each modality was preprocessed. We applied a band-pass filter so called difference of Gaussians (DoG), to emphasize the edges in addition to removing high and low frequency noise. The DoG filter which is the difference of two Gaussian kernels with different $\sigma$ is defined as follows:

\begin{equation}
\begin{split}
D(I, \sigma_0, \sigma_1) = [G(x,y,\sigma_0)- G(x,y,\sigma_1)] * I (x,y) \;,
\label{eq-7}
\end{split}
\end{equation}
\noindent where $D$ is the DoG filtered image, $*$ is the convolution operator, and $ G$ is the Gaussian kernel which is defined in: 
\begin{equation}
\begin{split}
G(x,y,\sigma) = \dfrac{1}{\sqrt{2\pi\sigma^2}}  e^ {-\dfrac{x^2+y^2}{2\sigma^2}}  \;.
\label{eq-8}
\end{split}
\end{equation}

To increase the number of training patterns and prevent overfitting each face image was partitioned into the $40\times 40$ cropped patches with stride of 10. The same preprocessing procedure (DoG filtering and $40\times 40$ cropping) was applied to the Polarimetric Thermal Face images dataset. Among the 60 total subjects in the dataset, 25 subjects were used to train the network and the remaining 35 subjects were used for testing. The training set was used to transform the visible and polarimetric thermal features to the common latent embedding subspace. For each modality there are four baseline and 12 expression images at three different ranges. After partitioning each image into $40\times 40$ patches, the genuine and impostor pairs were constructed. For each patch, we consider the corresponding location patch from another modality of the same subject but in different domain as the genuine pair. For the impostor pair, the same location patches from different modalities but from different subjects were selected. Therefore, the number of the generated impostor pairs were significantly larger than the genuine pairs. For the sake of balancing the training set, we considered the same number of genuine and impostor pair.     

\begin{figure}
	\begin{center}
		\includegraphics[width=1.2\linewidth]{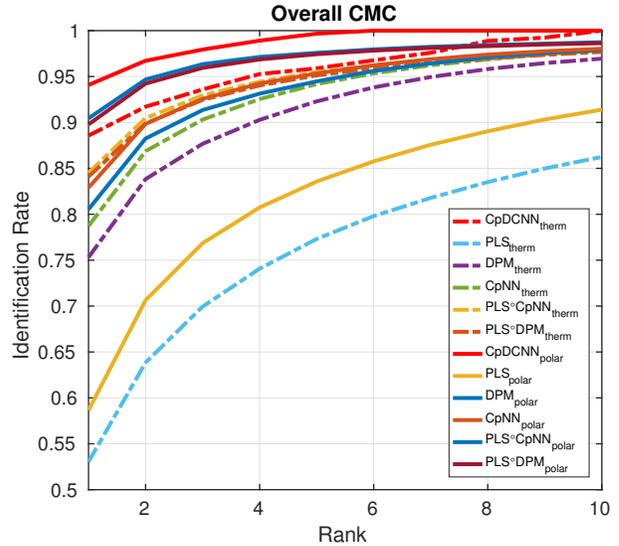}
		
	\end{center}
	\caption{Overall CMC curves from testing PLS, DPM, CpNN, and CpDCNN using polarimetric and thermal probe samples, matching against a visible spectrum gallery.}
	\label{fig:cmc}
\end{figure}

\begin{table*}
	\begin{center}
		\begin{tabular}{|c|c|c|c|c|c|c|c|}
			\hline
			Scenario & \multicolumn{7}{c|}{Rank-1 Identification Rate} \\ \cline{2-8}
			\multirow{3}{*}{} & Probe  & PLS  & DPM  & CpNN & PLS$\circ$DPM & PLS$\circ$CpNN & CpDCNN \\ \cline{2-8}  \hline  \hline
			Overall & Polar  &  0.5867 & 0.8054  & 0.8290 & 0.8979 &  0.9045& \textbf{0.9408} \\ \cline{2-8} 
			& Therm  & 0.5305  & 0.7531  & 0.7872 & 0.8409 &  0.8452 & \textbf{0.8857} \\ \hline
			\multirow{2}{*}{}Expressions & Polar  & 0.5658  &   0.8324& 0.8597  & 0.9565 & 0.9559  & \textbf{0.9637}  \\ \cline{2-8} 
			& Therm  & 0.6276  & 0.7887  & 0.8213 & 0.8898 &  0.8907 & \textbf{0.9124}  \\ \hline
			\multirow{2}{*}{} Range 1 Baseline & Polar  & 0.7410  & 0.9092   & 0.9207 & 0.9646 & 0.9646 & \textbf{0.9721} \\ \cline{2-8} 
			& Therm  & 0.6211  & 0.8778   & 0.9102  & 0.9417 & 0.9388  & \textbf{0.9534} \\ \hline
			\multirow{2}{*}{} Range 2 Baseline & Polar  & 0.5570  & 0.8229  & 0.8489 & 0.9105 & 0.9187 & \textbf{0.9317}  \\ \cline{2-8} 
			& Therm  & 0.5197  & 0.7532  & 0.7904 & 0.8578 &  0.8586 & \textbf{0.8868} \\ \hline
			\multirow{2}{*}{} Range 3 Baseline & Polar  & 0.3396  & 0.6033  & 0.6253 & 0.6445 & 0.6739 & \textbf{0.8346}  \\ \cline{2-8} 
			& Therm  & 0.3448  & 0.5219  & 0.5588 & 0.5768 & 0.6014  &\textbf{0.7754}   \\ \hline
		\end{tabular}
	\end{center}
	\caption{Rank-1 identification rate for cross-spectrum face recognition using polarimetric thermal and thermal probe imagery.}
	\label{table:table1}
\end{table*}

For the Polarimetric Thermal Face dataset, we considered the same CpDCNN architecture that was trained on the Notre Dame LWIR Face dataset. We passed $S_0$, $S_1$, and $S_2$ to the Pol-DCNN's three channels as the input . However, since, the Vis-DCNN and Pol-DCNN have already learned useful features of face in visible and LWIR modalities, and to prevent the problem of overfitting, all the layers of the network were not trained on this dataset. We conducted different experiments such that in each experiment different layers of the CpDCNN were trained and the remaining layers were fixed. The best results belong to the condition when the last three convolutional layers of each network (Vis-DCNN and Pol-DCNN) were trained and the remaining convolutional layers were frozen and kept the weights of the network which was trained on the Notre Dame LWIR Face dataset. In this way, the network extracts the thermal and visible low-level features (i.e.,edges and corners) to some extent in the first convolutional layers and learns the complementary information related to polarimetric thermal images ($S_1, S_2$) in the last three convolutional layers. It should be noted that we applied global pooling on the feature map related to the last convolutional layer. This tweak helped the network to reduce the number of parameters and keep the global information from the last feature maps. Reducing the number of parameters led us to overcome the problem of overfitting which was one of the main challenges in this work.

In each experiment the dataset was partitioned to train and test randomly. The experiment was repeated 100 times and the same set of train and test was used to evaluate PLS, DPM, CpNN, PLS$\circ$DPM, and PLS$\circ$CpNN and the proposed CpDCNN network, and the results were averaged over the 100 trials. Figure~\ref{fig:cmc} shows the overall cumulative matching characteristics (CMC) curves for our proposed method and the other state-of-the-art methods over all the three different ranges as well as the expressions data at Range 1. For the sake of comparison, in addition to polarimetric thermal-to-visible face recognition performance, Figure~\ref{fig:cmc} also shows the results for the conventional thermal-to-visible face recognition. In the conventional thermal-to-visible face recognition, all the mentioned methods exactly follow the same procedure as before, with only using $S_0$ modality. Figure~\ref{fig:cmc} shows that exploiting the polarization information of the thermal spectrum improves cross-spectrum face recognition performance compared to the conventional one. Figure~\ref{fig:cmc} also shows the superior performance of our approach compared to the state-of-the-art methods. In addition, our method could achieve prefect accuracy of 1 at Rank-6 and above. 

Table~\ref{table:table1} tabulates the Rank-1 identification rates for five different scenarios: overall (which corresponds to Figure~\ref{fig:cmc}), Range 1 expressions, Range 1 baseline, Range 2 baseline, and Range 3 baseline. In our framework, exploiting polarization information enhance the Rank-1 identification rate by 1.87\%, 5.13\%, 4.49\%, and 5.92\% for Range 1 baseline, Range 1 expression, Range 2 baseline, and Range 3 baseline compared to the conventional thermal-to-visible face recognition. This table shows the effectiveness of our method in utilizing polarization information to enhance the cross-spectrum face recognition problem. It also reveals that using deep coupled convolutional neural network techniques with contrastive loss function to transform different modalities into the distinctive common embedding space is superior to the other embedding techniques such as PLS$\circ$CpNN.       

\section{Conclusion}
We have introduced a novel approach to exploit polarimetric information for the purpose of thermal-to-visible face recognition. We have proposed to use coupled convolutional neural network  to learn deep global discriminative features. The proposed network is capable of transforming the visible and polarimetric thermal modalities into a common discriminative embedding space. We prevented overfitting by training the network on relatively large datasets and fine-tune it with the polarimetric dataset. We compared our method with state-of-the-art thermal to visible face recognition methods and showed the superiority of our method over them. The results also revealed that the polarimetric thermal information can be exploited to boost the conventional thermal-to-visible face recognition performance. 

\begin{center}
	ACKNOWLEDGEMENT
\end{center}

This work is based upon a work supported by the Center for Identification Technology Research (CITeR), an NSF Industry/University Cooperative Research Center (I/UCRC) funded under Grant CNS-$1650474$.

{\small
\bibliographystyle{ieee}
\bibliography{egbib}
}

\end{document}